# Can Ensembling Pre-processing Algorithms Lead to Better Machine Learning Fairness?

**Khaled Badran**
Concordia University

**Pierre-Olivier Côté**
Polytechnique Montréal

**Amanda Kolopanis**
Concordia University

**Rached Bouchoucha**
Polytechnique Montréal

**Antonio Collante**
Concordia University

**Diego Elias Costa**
Université du Québec à Montréal

**Emad Shihab**
Concordia University

**Foutse Khomh**
Polytechnique Montréal

*Abstract*—As machine learning (ML) systems get adopted in more critical areas, it has become increasingly crucial to address the bias that could occur in these systems. Several fairness pre-processing algorithms are available to alleviate implicit biases during model training. These algorithms employ different concepts of fairness, often leading to conflicting strategies with consequential trade-offs between fairness and accuracy. In this work, we evaluate three popular fairness pre-processing algorithms and investigate the potential for combining all algorithms into a more robust pre-processing ensemble. We report on lessons learned that can help practitioners better select fairness algorithms for their models.

■ **MACHINE LEARNING** (ML) systems are becoming further integrated into vital areas that impact our daily lives (e.g., finances, judiciary process, etc.). These systems can contain implicit biases, which generally consist of improper distributions between groups and characteristics. As such, many ML systems exhibit prejudice towards underrepresented groups due to imbalanced datasets, which impact the livelihood of many individuals.

Numerous examples of biased ML systems exist, each causing severe socio-economical consequences to their users. In one example, a financial institution's model refused several eligible mortgage loan applicants from minority groups because the model was trained on a dataset with embedded implications and imbalanced distribution [1]. Another instance is when Amazon's experimental job recruitment system primarily rejected women applicants technical positions [2] because the training data had an imbalance between the submitted resumes of past men and women employees [2]. In essence, these examples align with the growing body of research [3] that point to bias in the dataset as the core cause of ML bias.

In light of the potential consequences of an intelligent system's prejudice towards certain individuals, ML practitioners proposed approaches to address the bias in the dataset. One type of approach directly manipulates the training data before presenting it to the ML model (i.e., ***fairness pre-processing algorithms***). The objective of these algorithms is to promote *fairness* in a statistical sense, meaning that the algorithms try to transform the dataset to have an equal distributive representation of a specified protected attribute to prompt a fairer model. Since these algorithms target the dataset, ML practitioners can use them with any ML model (e.g., Random Forest) while also addressing the principal source of ML bias (the training dataset). However, manipulating the dataset can impact the model's accuracy, which may concern ML practitioners.

Each fairness pre-processing algorithm prioritizes different goals. For example, some algorithms mainly optimize for fairness [4], [5] whereas others have considerations for accuracy [6]. This results in 1) each algorithm offering its own fairness-accuracy trade-offs and 2) no single algorithm yielding optimal fairness and accuracy performance in every case [7]. Hence, ML practitioners face a challenge when selecting the best algorithm for their applications.

To address this challenge, we set out on two endeavors in this study. First, we evaluate the performance of several fairness pre-processing algorithms to highlight their strengths and weaknesses in fairness and accuracy. This allows practitioners to better understand the trade-offs of each fairness processing algorithm and make a more informed decision when selecting one for their ML systems.

Second, we build upon the intuition of Nina et al. [8], where they speculate that diversity in opinions leads to fairer treatment. Specifically, we investigate whether ensembling multiple fairness pre-processing algorithms can provide further improvements to the model's fairness and accuracy (i.e., overcome their shortcomings). The benefit of the ensemble approach is that it considers the fairness transformations offered by multiple fairness pre-processing algorithms when making predictions. Moreover, the ensemble approach will alleviate some of the pressures from ML practitioners when selecting the most suitable fairness pre-processing algorithms for their scenario.

Therefore, the contributions of this study are listed as follows:

- We provide practitioners with empirical evidence on the performance of three popular fairness pre-processing algorithms by highlighting their impact on the model's fairness and accuracy.
- We investigate whether an ensemble approach can offer practitioners more robust fairness and accuracy performance.
- We provide practitioners with actionable recommendations on which fairness pre-processing algorithms to select for their application.
- We share the implementation[1] of the ensemble approach with the community to enable more experimentation and accelerate future research into this area.

[1] https://github.com/KhaledBadran/FairBoost

## Evaluation Setup

Our study aims to evaluate the impact of fairness pre-processing algorithms and their ensemble on a model's fairness and accuracy. In the following, we describe how we achieved our goal by selecting real-world datasets, widely-used ML models, the metrics that assess fairness and accuracy, and our experiment setup.

### Fairness Pre-Processing Algorithms

We select three fairness pre-processing algorithms to evaluate based on two criteria. First, we utilize algorithms that have been widely studied by prior research as it highlights their popularity. Second, each algorithm applies a different transformation to the dataset to avoid biasing the ensemble towards one type of fairness transformation. Notably, we leverage the implementation of these algorithms provided from *AIF360*[2].

- **Reweighing (RW)**: Removes discriminatory aspects in the data by assigning weights to the training examples in each *(group, label)* combination in the dataset [9]. The algorithm considers group fairness by transforming the dataset to have more equity in positive outcomes on the protected attribute for the privileged and unprivileged groups.
- **Learning Fair Representations (LFR)**: Identifies potential biases amongst the features by finding a latent representation that encodes the data well but obscures information about the determined protected attributes [6]. The algorithm promotes group and individual fairness by abstracting relationships between attributes while aiming to maintain the model's predictive performance.
- **Optimized Pre-processing (OP)**: Learns a probabilistic transformation of the dataset by editing the features and labels with the consideration of group fairness, individual distortion, and data fidelity constraints [10]. The algorithm improves group fairness by automatically attempting to reduce the statistical parity difference between all possible combinations of groups in the dataset.

---
[2]https://aif360.mybluemix.net/

### Datasets

We select three datasets that include socially sensitive features extracted from real-world data. All datasets are used in prior work to evaluate fairness-enhancing techniques and represent a binary classification problem. More specifically, we use (1) the *German* credit dataset [11] which includes data on 1,000 bank account holders and their credit evaluation (e.g., good or bad). (2) The *COMPAS* dataset [12] has information on 10,000 criminal defendants and was used by courts to help judges and parole officers make better decisions. (3) The *Adult* Census Income dataset [13] is used to determine whether a person earns more than $50,000 USD and includes 48,842 instances. Finally, we note that many fairness pre-processing algorithms, including two that are evaluated in this study (i.e., *LFR* [6] and *RW* [9]), work only on structured data. Therefore, we only use structured datasets in our paper to ensure that all the fairness pre-processing algorithms can be applied to these datasets.

In all datasets in this study, we consider 'sex' as the protected attribute that will be addressed by the fairness pre-processing algorithms. The reason for selecting 'sex' as the protected attribute is twofold. First, some of the fairness pre-processing algorithms (e.g., *RW* [1]) work only with binary attributes. Among all attributes in the three datasets used in the evaluation, 'sex' is the only protected attribute that is binary. Second, 'sex' is the only protected attribute that is mutual across all datasets. Therefore, selecting it makes our evaluation consistent across the datasets.

### Models

We select *Logistic Regression* and *Random Forest* for our evaluation. Both models have been subject to previous studies on fairness algorithms [7]. Thus, our study complements the previous studies by evaluating the fairness of the selected algorithms using an ensemble approach. Furthermore, the models have different characteristics [14]. *Logistic Regression* is effective in linear problems, whereas *Random Forest* is robust to deal also with non-linear classification. As well, these classifiers have different assumptions about the analyzed data and can deal with overfitting [15].

We select Logistic Regression and Random

Table 1: Impact of fairness pre-processing algorithms on the normalized disparate impact and accuracy of both models. The baseline reports the performance of each model with no fairness pre-processing algorithms. White cells (☐) indicate no statistical differences against the baseline, while green (☐) and red cells (☐) highlight the experiments where fairness pre-processing algorithms respectively improved or worsened the original model's performance.

|  |  | Random Forest Classifier | | | |  | Logistic Regression Classifier | | | |
| --- | --- | --- | --- | --- | --- | --- | --- | --- | --- | --- |
|  | Dataset | Baseline | LFR | OP | RW |  | Dataset | Baseline | LFR | OP | RW |
| **Fairness** | German | 0.821 | 0.856 | 0.821 | 0.873 | **Fairness** | German | 0.860 | 0.862 | 0.863 | 0.942 |
|  | COMPAS | 0.749 | 0.976 | 0.869 | 0.900 |  | COMPAS | 0.690 | 0.963 | 0.784 | 0.863 |
|  | Adult | 0.000 | 0.887 | 0.788 | 0.732 |  | Adult | 0.000 | 0.891 | 0.424 | 0.686 |
| **Accuracy** | German | 0.798 | 0.812 | 0.796 | 0.796 | **Accuracy** | German | 0.806 | 0.814 | 0.804 | 0.805 |
|  | COMPAS | 0.691 | 0.669 | 0.698 | 0.688 |  | COMPAS | 0.701 | 0.686 | 0.708 | 0.697 |
|  | Adult | 0.451 | 0.404 | 0.423 | 0.412 |  | Adult | 0.484 | 0.404 | 0.491 | 0.495 |

Forest in our study to evaluate the impact of fairness pre-processing algorithms on two popular ML models. Both models have been subject to previous studies on fairness algorithms [7]. Also, each of these classifiers has different assumptions about the training data [15], and different characteristics [14]. Logistic Regression is effective in linear problems, whereas Random Forest is robust with non-linear classification. Finally, in our experiment, we utilize the *Scikit-Learn*[3] implementation for both models with their default parameters.

Metrics

To assess a model's fairness and accuracy, we leverage the following metrics:

- **Normalized Disparate Impact (NDI)**: We select *Disparate Impact* as the measure of fairness in our study. Although fairness can be defined using other metrics or notions, *Disparate Impact* is a widely popular measure of group fairness and is regarded as the most correlated with other fairness metrics in prior work [16], [17]. On its own, disparate impact (*DI*) is the probability ratio between the positive outcomes for unprivileged and privileged groups:

$$DI = \frac{P(Y=1)|P(D=unprivileged)}{P(Y=1)|P(D=privileged)}$$

However, this value can be more than 1.0 when the unprivileged group is assigned more positive outcomes (bias against the privileged group). Therefore, to unify how bias towards privileged or unprivileged groups is measured, we normalize the DI metric:

$$NDI = \begin{cases} DI & DI \leq 1.0 \\ DI^{-1} & DI > 1.0 \end{cases}$$

where the value of DI will always be between 0.0 (least fair representation) and 1.0 (maximum fairness).

- **F1 Score**: measures the harmonic mean between the precision and recall. This metric calculates the value as follows, where *TP* represents the true positive outcomes, *FP* depicts the false positive results, and *FN* denotes the false negative events:

$$F1 = \frac{TP}{TP + \frac{1}{2}(FP + FN)}$$

Experiment Setup

We aim to assess the performance of fairness pre-processing algorithms. Hence, we first establish a **baseline** consisting of a model's performance when trained on the original dataset. We follow the same procedure in all experiments. First, we randomly split the dataset into training and test sets, using a 70%-30% split. Depending on the experiments, we employ the fairness pre-processing algorithms to transform the training data and train our ML models with the newly created training set. We then evaluate the performance of the trained model using the held-out test set, assessing both its fairness and accuracy.

---
[3] https://scikit-learn.org/1.0

This process repeats ten times with random seeds to ensure that the results are not biased towards a specific train-test split. Finally, to assess whether the difference in results is statistically significant, we use the Mann-Whitney U test at the 5% level of significance (i.e., $\alpha = 0.05$).

### Fairness Pre-processing Algorithms Results

Table 1 presents the fairness and accuracy performance of the models without any pre-processing algorithm (baseline) and the models' performance when training data is transformed by the fairness pre-processing algorithms.

Overall, we observe that **fairness pre-processing algorithms significantly improve the model's fairness**. All fairness pre-processing algorithms improve the fairness metrics by at least 10% in both *COMPAS* and *Adult* datasets. Particularly in the case of the *Adult* dataset, the model went from completely biased (NDI = 0) to significantly less biased (NDI $\geq$ 0.42).

**Improvement of the model's fairness varies significantly across fairness pre-processing algorithms.** In terms of gain in fairness metrics, the *LFR* algorithm yields the best results, showing the highest fairness values in most of our experiments. The second-best algorithm was *RW*, which is the only algorithm to show a significant improvement on model's fairness in the *German* dataset (Logistic Regression), but overall ranks worse than *LFR* in fairness. Finally, *OP* ranks consistently in third place across all experiments.

**The algorithm with the best fairness improvement yields the worst impact on the models' accuracy.** When we look at the bottom half of Table 1, we notice that *LFR* was the only algorithm that exhibited a negative impact on the model's accuracy (red cells). Neither *RW* nor *OP* impacted the model's accuracy negatively, and in the scenarios run with *Logistic Regression*, we notice improvements from both algorithms in the *Adult* dataset.

Overall, while *LFR* performs the best in improving the models' fairness, it negatively impacted the models' accuracy in some evaluation scenarios, exhibiting a clearer fairness accuracy trade-off. *RW* and *OP* showed better stability by improving the fairness of the models without compromising their accuracy. The variety of trade-offs offered by each algorithm can be a reflection of the different optimization goals (e.g., accuracy) in their implementations.

### The Ensemble Approach

So far, we see that fairness pre-processing algorithms provide different trade-offs, which raises challenges for practitioners to select the best algorithm for their ML systems. However, insights from prior work can help guide the way towards a solution. Nina et al. [8] intuited that considering more diverse viewpoints results in more equitable treatment. Following that intuition, we investigate whether an approach that combines the different perspectives of multiple fairness pre-processing algorithms via an ensemble can lead to better fairness and accuracy performance. In this section, we describe the architecture of this ensemble approach.

Figure 1 provides an overview of the ensemble approach. First, we apply the three pre-processing algorithms $FP_X$ on the original dataset $D$ to create multiple (transformed) datasets $D'_X$. Then, the models learn from the transformed datasets. Finally, the models perform majority voting to make predictions.

### Ensemble Approach Results

Table 2 shows the results of the model's performance when using the ensemble methods. The baseline for this experiment is the model's performance using individual fairness pre-processing algorithms, i.e., the last three columns of Table 1.

In most cases, **combining fairness pre-processing algorithms does not improve upon the individual pre-processing algorithm's impact on fairness and accuracy**. This is particularly true in the experiments with *Random Forest*. Combining *LFR* with other algorithms did not alleviate its negative impact on accuracy but caused the ensemble to worsen its original fairness improvement. The ensemble methods also consistently yielded worse accuracy on *RW* and *OP*, which initially had no negative impact on the model's accuracy (see Table 1).

We observe **some improvements when combining *OP* with the other two algorithms in the *Logistic Regression's fairness***. The original fairness metric of the model using the *OP* algorithm in the *Adult* dataset increased from 0.42

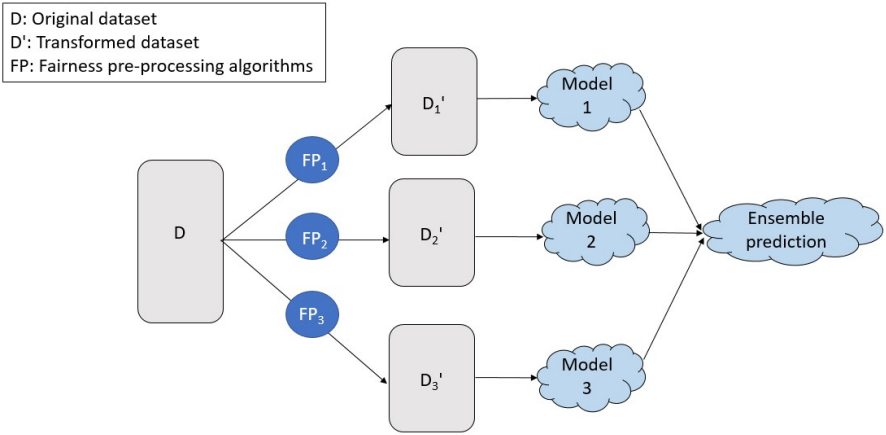

Figure 1: An Illustration of the Ensemble Approach

Table 2: Performance comparison of fairness pre-processing algorithms against the different ensembles. Each table compares the performance of an individual algorithm (*LFR, RW, OP*) versus the ensemble methods. We employed a similar code scheme to highlight that the ensemble showed improvement (□), same performance (□), or worsened (□) the performance of the model with individual fairness algorithms.

(a) LFR with Random Forest

| Performance | Dataset | LFR | +OP | +RW | All |
|---|---|---|---|---|---|
| Fairness | German | 0.856 | 0.752 | 0.817 | 0.736 |
|  | COMPAS | 0.976 | 0.853 | 0.872 | 0.827 |
|  | Adult | 0.887 | 0.608 | 0.839 | 0.540 |
| Accuracy | German | 0.812 | 0.793 | 0.788 | 0.780 |
|  | COMPAS | 0.669 | 0.653 | 0.654 | 0.643 |
|  | Adult | 0.404 | 0.330 | 0.381 | 0.326 |

(b) LFR with Logistic Regression

| Performance | Dataset | LFR | +OP | +RW | All |
|---|---|---|---|---|---|
| Fairness | German | 0.862 | 0.842 | 0.895 | 0.862 |
|  | COMPAS | 0.963 | 0.784 | 0.851 | 0.865 |
|  | Adult | 0.891 | 0.697 | 0.891 | 0.697 |
| Accuracy | German | 0.814 | 0.805 | 0.803 | 0.802 |
|  | COMPAS | 0.686 | 0.690 | 0.682 | 0.678 |
|  | Adult | 0.404 | 0.389 | 0.403 | 0.389 |

(c) RW with Random Forest

| Performance | Dataset | RW | +LFR | +OP | All |
|---|---|---|---|---|---|
| Fairness | German | 0.873 | 0.817 | 0.819 | 0.736 |
|  | COMPAS | 0.900 | 0.872 | 0.927 | 0.827 |
|  | Adult | 0.732 | 0.839 | 0.692 | 0.540 |
| Accuracy | German | 0.796 | 0.788 | 0.785 | 0.780 |
|  | COMPAS | 0.688 | 0.654 | 0.682 | 0.643 |
|  | Adult | 0.412 | 0.381 | 0.378 | 0.326 |

(d) RW with Logistic Regression

| Performance | Dataset | RW | +LFR | +OP | All |
|---|---|---|---|---|---|
| Fairness | German | 0.942 | 0.895 | 0.878 | 0.862 |
|  | COMPAS | 0.863 | 0.851 | 0.865 | 0.865 |
|  | Adult | 0.686 | 0.891 | 0.433 | 0.697 |
| Accuracy | German | 0.805 | 0.803 | 0.800 | 0.802 |
|  | COMPAS | 0.697 | 0.682 | 0.696 | 0.678 |
|  | Adult | 0.495 | 0.403 | 0.475 | 0.389 |

(e) OP with Random Forest

| Performance | Dataset | OP | +LFR | +RW | All |
|---|---|---|---|---|---|
| Fairness | German | 0.821 | 0.752 | 0.819 | 0.736 |
|  | COMPAS | 0.869 | 0.853 | 0.927 | 0.827 |
|  | Adult | 0.788 | 0.608 | 0.692 | 0.540 |
| Accuracy | German | 0.796 | 0.793 | 0.785 | 0.780 |
|  | COMPAS | 0.698 | 0.653 | 0.682 | 0.643 |
|  | Adult | 0.423 | 0.330 | 0.378 | 0.326 |

(f) OP with Logistic Regression

| Performance | Dataset | OP | +LFR | +RW | All |
|---|---|---|---|---|---|
| Fairness | German | 0.863 | 0.842 | 0.878 | 0.862 |
|  | COMPAS | 0.784 | 0.784 | 0.865 | 0.865 |
|  | Adult | 0.424 | 0.697 | 0.433 | 0.697 |
| Accuracy | German | 0.804 | 0.805 | 0.800 | 0.802 |
|  | COMPAS | 0.708 | 0.690 | 0.696 | 0.678 |
|  | Adult | 0.491 | 0.389 | 0.475 | 0.389 |

to 0.69 when combined with the *LFR* algorithm. Similarly, combining *OP* with *RW* improved the model's original fairness from 0.78 to 0.86 in the *COMPAS* dataset. In both cases, the improvement in fairness produced worse accuracy, particularly on the *Adult* dataset, where the *Logistic Regression* model's accuracy fell to 0.38 (lower than the original model with no pre-processing algorithms).

Overall, while we identify trade-offs in the original fairness pre-processing algorithms, our results with the ensemble approach show that combining algorithms may not improve a model's fairness and accuracy. Next, we report on important lessons we learned throughout this study.

## Lessons Learned

We faced several challenges when conducting our evaluation, and we report the lessons learned for practitioners here.

**Improving models' fairness with ensembles is not straightforward.** While it is common amongst practitioners to opt for an ensembling approach to improve the model's accuracy, our experiments showed that successfully using ensembles to improve the model's fairness is challenging. The ensembles exhibit the worst accuracy results in comparison to the models by themselves (i.e. leveraging only one algorithm at a time), which is the opposite behavior of what we could expect. By having a closer look at the results from Table 2, a trend emerges. When increasing the heterogeneity within an ensemble, we can generally observe drops in fairness and accuracy performances. This concept fosters from our observations of the results. When the number of algorithms used in the ensemble increase, the performance generally decreases, especially in terms of accuracy. This behavior is particularly visible in Table 2a, where the ensemble with three fairness algorithms consistently performs worse than the ones created with pair-wise combination. **For these reasons, we recommend that practitioners keep an eye on the factors contributing to the diversity of the ensemble (e.g., number of algorithms, bootstrapping, etc.) as it can impact the performance of the model.**

**Practitioners should carefully select the hyper-parameters of their fairness pre-processing algorithms.** We found some of the fairness pre-processing algorithms to be extremely sensitive in selecting hyper-parameter values and producing invalid datasets (with null values) if parameters were not properly set. For example, *LFR* generated unstable results in several instances because the hyper-parameter values needed various adjustments for the corresponding input dataset. Practitioners have to carefully tune the algorithms for their particular dataset, not only to achieve the best result but to prevent errors in the ML training pipeline. We recommend that practitioners optimize the hyper-parameters of the fairness pre-processing algorithms using a search-based technique over a parameter space where the goal can be to optimize for both the fairness and accuracy of the model.

**Fairness pre-processing algorithms require very specific data formats.** We found the available open-source[4] implementation of the fairness pre-processing algorithms to be inflexible towards different data formats. The *OP* implementation requires one-hot of the dataset to function, and the *LFR* implementation only permits binary labels with the values of 0 and 1. For instance, the labels in the *German* dataset are values of 1 and 2, however, to use the *LFR* implementation we update them to 0 and 1. **Practitioners should carefully analyze the data format requirements of each algorithm's implementation before adopting any algorithm.**

**Different ensembling techniques have different impact on model fairness and accuracy.** In The Ensemble Approach section, we present a majority voting technique to ensemble fairness pre-processing algorithms. Nonetheless, to generalize our results, we evaluated the impact of using bagging (majority voting with bootstrapping) and stacking ensemble techniques on the performance of the model. The results [5] show that both bagging and stacking ensemble techniques achieve lower fairness performance overall compared to the individual fairness pre-processing algorithms. On the other hand, stacking achieves better accuracy in half of the cases (29/54) compared to the individual fairness pre-processing algorithm, while bagging archives poorer accuracy in all cases. **Hence, we find that ensembling fairness pre-processing algorithms does not improve the model fairness overall, regardless of the used ensembling technique.**

---
[4] https://github.com/Trusted-AI/AIF360
[5] https://zenodo.org/record/7259845

## Threats to validity

During our experiments, the *LFR* algorithm did not work with our selected datasets under its default parameter values. The LFR generated invalid datasets, e.g., including one label value, wrongly transforming the feature values. Hence, we had to adjust the *LFR* hyper-parameters (following the recommendations of the original study) to a set of values that yielded a usable pre-processed dataset for our experiments. While we did not perform hyper-parameter tuning, changing *LFR* parameters could put the *LFR* algorithm at an advantage compared to the other methods, creating a threat to the internal validity of the study. Another threat to the external validity of our study is that we included only three datasets. While these datasets are commonly used by related work and represent well real-case scenarios, our results may not generalize beyond the selected scenarios.

## Future Avenues

Our study sets the path for further investigation in this field. Future work can leverage the flexibility of ensembles and investigate the effectiveness of combining fairness pre-processing algorithms with fairness in-processing or post-processing algorithms. As well, other combinations of fairness pre-processing algorithms (e.g., by adding the Disparate Impact Remover algorithm) can be used to evaluate the ensemble approach. Lastly, further experimentation can explore the effect of using the ensemble approach on more complex classifiers such as Neural Networks.

## ■ REFERENCES

**Khaled Badran** is a Master's student at Concordia University, Montreal, Canada. His research interests are in software bots/chatbots, natural language understanding, and machine learning fairness. Contact him at k_badran@encs.concordia.ca.

**Pierre-Olivier Côté** is a Master's student at Polytechnique Montréal, Canada. His research interests are in quality assurance for machine learning systems, data cleaning, and machine learning fairness. Contact him at pierre-olivier.cote@polymtl.ca.

**Amanda Kolopanis** is a Master's student at Concordia University, Montreal, Canada. Her research interests target incorporating gender and society in artificial intelligence, machine learning, and natural language understanding. Contact her at amanda.kolopanis@mail.concordia.ca.

**Rached Bouchoucha** is a student in Polytechnique Montreal, Canada. His main research interests are on the quality assurance, testing, and maintenance of deep learning models, with a big focus on reinforcement learning techniques. Email him at rached.bouchoucha@polymtl.ca.

**Antonio Collante** is a PhD student at Concordia University, Montreal, Canada. His research interests are in software project management, software analytics, mining software repositories, and machine learning fairness. Contact him at a_collan@encs.concordia.ca.

**Diego Elias Costa** is an Assistant Professor in the Department of Computer Science at the Université du Québec à Montréal (UQAM), Canada. His research interests cover various software engineering related topics such as dependency management, SE4AI, performance testing, and software engineering bots. You can find more about him at https://diegoeliascosta.github.io/.

**Emad Shihab** is a full professor and Concordia research chair in the Department of Computer Science and Software Engineering at Concordia University. Dr. Shihab leads the Data-driven Analysis of Software (DAS) lab. He received his PhD from Queens University. Dr. Shihab's research interests are in Software Quality Assurance, Mining Software Repositories, Technical Debt, Mobile Applications and Software Architecture. He worked as a software research intern at Research In Motion in Waterloo, Ontario and Microsoft Research in Redmond, Washington. Dr. Shihab is a member of the IEEE and ACM. More information can be found at http://das.encs.concordia.ca.

**Foutse Khomh** is a Full Professor of Software Engineering at Polytechnique Montréal, a Canada CIFAR AI Chair on Trustworthy Machine Learning Software Systems, and a FRQ-IVADO Research Chair on Software Quality Assurance for Machine Learning Applications. He is also a member of Mila - Quebec AI Institute. Contact him at foutse.khomh@polymtl.ca.